\documentclass{article}
\usepackage{spconf,amsmath,graphicx, multirow}
\usepackage{caption}
\usepackage{subcaption}
\usepackage{xcolor}
\usepackage{amsfonts}

\DeclareMathOperator*{\argmax}{arg\,max}

\title{I see what you hear: a vision-inspired method to localize words}
\name{\small Mohammad Samragh, Arnav Kundu, Ting-Yao Hu, Aman Chadha*, Ashish Srivastava*\thanks{* contributed when employed by Apple}, Minsik Cho, Oncel Tuzel, Devang Naik}
\address{Apple}
\begin{document}
%
\maketitle
\begin{abstract}
This paper explores the possibility of using visual object detection techniques for word localization in speech data. Object detection has been thoroughly studied in the contemporary literature for visual data. Noting that an audio can be interpreted as a 1-dimensional image, object localization techniques can be fundamentally useful for word localization. Building upon this idea, we propose a lightweight solution for word detection and localization. We use bounding box regression for word localization, which enables our model to detect the occurrence, offset, and duration of keywords in a given audio stream. 
We experiment with LibriSpeech and train a model to localize 1000 words. Compared to  existing work~\cite{segal2019speechyolo}, our method reduces model size by $94\%$, and improves the F1 score by $6.5\%$.
\end{abstract}
\setlength{\abovedisplayskip}{3pt}
\setlength{\belowdisplayskip}{3pt}

\vspace{-0.25cm}
\section{Introduction}
\vspace{-0.25cm}
Recent advancements in automatic speech recognition (ASR) technologies have made human machine interaction seamless and natural~\cite{zhang2020pushing, chung2021w2v, xu2021self}. ASR systems are designed to accurately comprehend any speech data, hence, they are often computationally expensive, power hungry, and memory intensive. Acoustic models with limited-vocabulary can only recognize certain words, but they offer computationally efficient solutions~\cite{rybakov2020streaming, alvarez2019end,tang2018deep}. In this paper we focus on the latter.

Limited-vocabulary models are important for several reasons. First, users often interact with their devices using simple commands~\cite{tang2022temporal}, and recognizing these commands may not necessarily require an ASR model. Second, limited-vocabulary models are needed to detect trigger phrases, e.g., ``Alexa, OK Google, hey Siri'', which indicate that a user wants to interact with the ASR model.
A limited vocabulary model should 
\textbf{(a)}~accurately recognize if certain words are spoken, and \textbf{(b)}~precisely locate the occurrence time of the words. The latter is rather important due to privacy and efficiency reasons, as an ASR model should only be queried/executed when users intend to interact with it. Additionally, a limited-vocabulary model with localization capabilities can improve the accuracy of an ASR model by providing noise-free payloads to it.


A plethora of existing work focus on keyword detection~\cite{rybakov2020streaming, alvarez2019end,tang2018deep}, yet, efficient and accurate word localization needs more investigation. 
In computer vision, object localization has been solved using bounding box detection~\cite{yolov3, liu2016ssd,zhou2019objects}. Since an audio can be interpreted as a 1-D image, similar techniques can be used in principle to localize words in an audio. The first effort in this track is SpeechYolo~\cite{segal2019speechyolo}, which shows the great potential of using visual object detection techniques for word localization.

This paper presents an alternative vision-based word localizer. In our design, we pay attention to several important properties: \textbf{(a)}~having small memory footprint, \textbf{(b)}~the ability to process streaming audio, \textbf{(c)}~accurate word detection and \textbf{(d)}~accurate word localization. To achieve these goals, we propose customized metrics that indicate the presence or absence of words in the input audio. We then devise appropriate loss functions and train our model to perform three tasks simultaneously on streaming audio: detection, classification, and localization. We experiment with LibriSpeech and train a model to localize 1000 words. Compared to SpeechYolo, our model is $94\%$ smaller, is capable of processing streaming audio, and achieves $6.5\%$ better F1 score.

\vspace{-0.25cm}
\section{Related work}
\vspace{-0.25cm}
The task of detecting limited vocabulary has been studied in the keyword spotting literature~\cite{rybakov2020streaming, alvarez2019end,tang2018deep},
where the goal is to detect if a keyword is uttered in a segmented audio. Our problem definition is more challenging as we expect our model to precisely localize keywords in a streaming audio. In addition, the majority of existing literature target a small vocabulary, e.g., 12-36 words in Google Speech Commands~\cite{warden2018speech}, whereas we show scalability to 1000 words in our experiments.

Post processing approaches can be utilized for word localization. In DNN-HMM models~\cite{shrivastava2021optimize}, for instance, the sequence of predicted phonemes can be traced to find word boundaries. Another example is~\cite{palaz2016jointly} where word scores are generated by a CNN model for streaming audio segments, and the location of the words can be estimated accordingly. 
Since the above models are not directly trained for localization, their localization performance is likely not optimal. A more precise localization can be obtained by forced alignment after transcribing the audio using an ASR model~\cite{forcedalignment}, or by coupling an ASR model with a CTC decoder~\cite{ctc}. However, these solutions are computationally expensive.

Recently, principles from Yolo object detection~\cite{yolov3} have been used for speech processing, which incorporate both detection and accurate localization into the model training phase. YOHO~\cite{venkatesh2022you} can localize audio categories, e.g., it can distinguish music from speech. A more related problem definition to ours is studied in SpeechYolo~\cite{segal2019speechyolo}, which  shows great potential of vision-based localization techniques by localizing words in segments of 1-second audio. In this paper, we present a more carefully designed word localizer that is capable of processing streaming audio. Additionally, we show that our design yields better detection and localization accuracy with smaller memory footprint.

\section{Problem formulation}
We aim to detect words in the lexicon of $c$-words, indexed by $w\in\{1, \dots, c\}$. 
Let $\{(w_1, b_1, e_1), \dots, (w_k, b_k, e_k) \}$ be ground-truth events in an utterance $X\in \mathbb{R}^T$. Each event $(w, b, e)$ contains the word label $w\in\{1\dots c\}$, the event beginning time $b$, and the event ending time $e$. Our goal is to train $f$ parameterized by $\theta$ that predicts proposals: 
\begin{equation}
    f(\theta, X) = \{(\hat{w}_k, \hat{b}_k, \hat{e}_i)\}_{i=1}^{k'}.
\end{equation}
The recognizer model $f(\theta, X)$ should be trained such that the $k'$ predicted events match the $k$ ground truth events. 


\section{Methodology}
\begin{figure}
    \centering
    \includegraphics[width=\columnwidth]{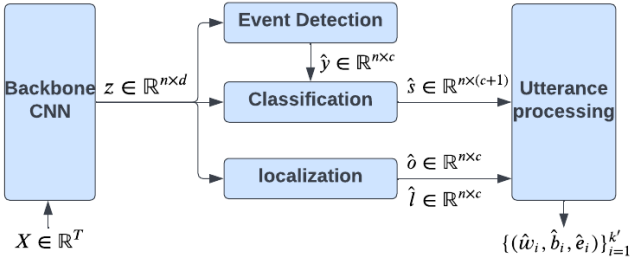}
    \caption{high-level overview of our word localization model.}
    \label{fig:overall}
    \vspace{-0.5cm}
\end{figure}

The overall flow of our word localization is shown in Figure~\ref{fig:overall}. The CNN backbone converts the input audio into a feature matrix $z$. The rest of the modules use this feature matrix to detect events (encoded by $\hat{y}$), classify event types (encoded by $\hat{s}$), and predict the event offset $\hat{o}$ and length $\hat{l}$. Finally, ($\hat{s}, \hat{o}, \hat{l}$) are processed by the utterance processing module and events are proposed as $\{(\hat{w}_i, \hat{b}_i, \hat{e}_i)\}_{i=1}^{k'}$.

\noindent{\bf Backbone.} The CNN model receives an utterance $X\in \mathbb{R}^T$ and converts it into a feature matrix $z\in \mathbb{R}^{n\times d}$. 
The rows of $z$ correspond to utterance segments $X[tS:tS+R]$, which we denote by $X_t$ in the remainder of the paper for notation simplicity. Here, $R$ is the length of each segment (a.k.a., the network's receptive field) and $S$ is the shift between two consecutive segments (a.k.a. the network's stride). In an utterance of length $T$, there are $n=\lfloor\frac{T-R}{S}\rfloor+1$ total segments.

\begin{figure}
    \centering
    \includegraphics[width=0.8\columnwidth]{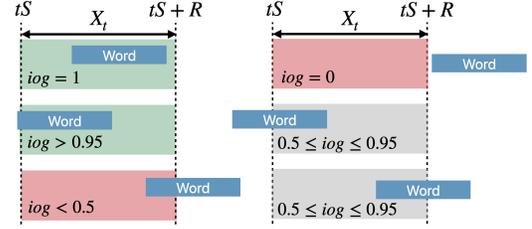}
    \caption{Examples of positive (green highlight), negative (red highlight), and ``don't care'' samples (grey highlight).}
    \label{fig:iog}
    \vspace{-0.5cm}
\end{figure}

\noindent{\bf Event detection.} The ground-truth event detection label is a binary matrix $y_{n\times c}$, where $y_{t,w}$ specifies whether $X_t$ contains the $w$-th word in the lexicon. To assign these hard-labels, we compute the intersection over ground-truth (iog) metric. Let $(w, b, e)$ be a ground-truth event. We compute:
\begin{equation}
    iog_{t,w} = \frac{\text{overlap [$(tS, tS+R)$,  $(b, e)$]}}{e-b}
\end{equation}
We threshold the $iog$ metric and assign labels accordingly:
\begin{itemize}
\setlength\itemsep{0em}
    \item if $iog_{t,w}>0.95$, word $w$ is almost perfectly contained in $X_t$; In this case we have $y_{t,w}=1$. 
    \item if $iog_{t,w}<0.5$, word $w$ is not contained in $X_t$; In this case we have $y_{t,w}=0$.  
    \item if $0.5\leq iog_{t,w} \leq 0.95$, word $w$ is partially contained in and partially outside $X_t$. The classification label is ``don't care'' in this case.
\end{itemize}
Figure~\ref{fig:iog} illustrates several examples for the above cases.

\noindent{\bf Event detection loss}.  We compute event probabilities as $\hat{y}_{n\times c}=sigmoid(W^{detection}\cdot{z})$. Our goal here is to encode the presence of word $w$ in $X_t$ as $\hat{y}_{t,w}$. If $X_t$ contains word $w$, $\hat{y}_{t,w}$ should be close to 1. To enforce this behaviour, we define the positive loss:
\begin{equation}
    L_{pos}(y, \hat{y}) = \frac{\sum\limits_{t=1}^{n}\sum\limits_{w=1}^{c} BCE(y_{t,w}, \hat{y}_{t,w})\cdot I(y_{t,w}, 1)}{\sum\limits_{t=1}^{n}\sum\limits_{w=1}^{c} I (y_{t,w}, 1)}
\end{equation}
where the numerator is the sum of binary cross entropy (BCE) loss over all elements with a ground truth label of 1, and the denominator is a normalizer that counts the number of positive labels. 
When word $w$ is not in $X_t$, $\hat{y}_{t, w}$ should be close to 0. To enforce this behaviour, we define the negative loss:
\begin{equation}
    L_{neg}(y, \hat{y}) = \frac{\sum\limits_{t=1}^{n}\sum\limits_{w=1}^{c} BCE(y_{t,w}, \hat{y}_{t,w})\cdot I(y_{t,w}, 0)}{\sum\limits_{t=1}^{n}\sum\limits_{w=1}^{c} I (y_{t,w}, 0)}
\end{equation}


\noindent{\bf Localization loss.} to predict event begin and end times, we adopt the CenterNet approach from visual object detection literature~\cite{zhou2019objects}. Let $c_t=\frac{2tS+R}{2S}$ be the center of $X_t$, and $(b,e)$ be the beginning and end of an event. We define the offset of the event as $o = \frac{b+e}{2S}-c_t$, and the length of the event as $l=\frac{e-b}{R}$. If our model can predict $o$ and $l$ accurately, we can calculate $b$, $e$. 
We generate offset prediction $\hat{o}_{n \times c} = W^{offset}\cdot z$ and length prediction $\hat{l}_{n \times c} = W^{length}\cdot z$. During training we minimize the normalized L1 distance between the ground-truth and predicted values. Equation~\ref{eq:offset_loss} shows the offset loss function $L_{o}$. The length loss function $L_l$ is defined similarly. 
\begin{equation}\label{eq:offset_loss}
\begin{aligned}
 L_{o}(o, \hat{o}, y) = \frac{\sum\limits_{t=1}^{n}\sum\limits_{w=1}^{c} |o_{t,w}- \hat{o}_{t,w}|\cdot I(y_{t,w}, 1)}{\sum\limits_{t=1}^{n}\sum\limits_{w=1}^{c} I (y_{t,w}, 1)} \\
\end{aligned}
\end{equation}
The predictions $(\hat{y}, \hat{o}, \hat{l})$ defined up to here can be used to generate region proposals for possible events. We initially applied non-maximum suppression (NMS), a technique widely used in image object localization~\cite{neubeck2006efficient}, to select the best non-overlapping proposals but many proposals were wrong. Our investigations unveiled two reasons for this matter:
\begin{itemize}
\setlength\itemsep{0em}
    \item \textbf{collision:} $X_t$ may contain multiple words, thus, $\hat{y}_{t,w}$ might be non-zero for multiple $w$. This makes it unlikely for NMS to make a correct proposal.
    \item \textbf{confusion:} even if there is only a single event within $X_t$, it might get confused, e.g., our model might confuse a ground-truth word ``two'' with ``too'' or ``to''.
\end{itemize}

\noindent{\bf Classifier.} To address the collision and confusion issues stated above, we propose to train a classifier that predicts: 
\begin{equation}\label{eq:softmax}
    \hat{s}_{n\times{c+1}} = softmax([W^{classifier}\cdot z]\otimes mask(\hat{y}))
\end{equation}
Here, $\hat{s}_{t,w}$ predicts the probability that the $X_t$: \textbf{(a)} contains the $w$-th word in the lexicon, for $w\leq c$, or \textbf{(b)} does not contain any of the words in the lexicon, for $w=c+1$. The $\otimes$ operator in Equation~\ref{eq:softmax} denotes element-wise multiplication, and $mask(\hat{y})$ is a binary tensor:
\begin{equation}
    mask_{t,w} = \begin{cases}
    1\quad if\ \  \hat{y}_{t,w}\geq 0.5\ \  or\ \  w=c+1\\
    0\quad  otherwise
    \end{cases}
\end{equation}
With the above masking formulation, our Softmax layer does not need to solve a $c+1$ classification problem. Instead, it solves the problem only for the positive classes proposed by $\hat{y}$  and the negative class. We train the softmax classifier using the cross-entropy loss:
\begin{equation}
        L_{s}(s, \hat{s}) = \frac{1}{n}\sum\limits_{t=1}^{n} CE(s_{t}, \hat{s}_{t})
\end{equation}
where $s$ is the ground-truth label and $CE$ is the cross-entropy loss. The ground truth $s_t \in \mathbb{R}^{c+1}$ is a one-hot vector that indexes either one of the $c$ word classes or the negative class.  In cases that $X_t$ contains more than one ground-truth event, $s_t$ indexes the event with smallest offset $o$. In essence, the Softmax classifier either rejects the events suggested by $\hat{y}$ or selects one event (which has maximum probability) from them. 

\noindent{\bf Total loss}. The training loss is given by:
\begin{equation}
    L = L_{pos} + L_{neg} + L_o + L_l + L_s
\end{equation}

\begin{figure}
    \centering
    \begin{subfigure}[b]{0.45\columnwidth}
        \includegraphics[width=0.8\columnwidth]{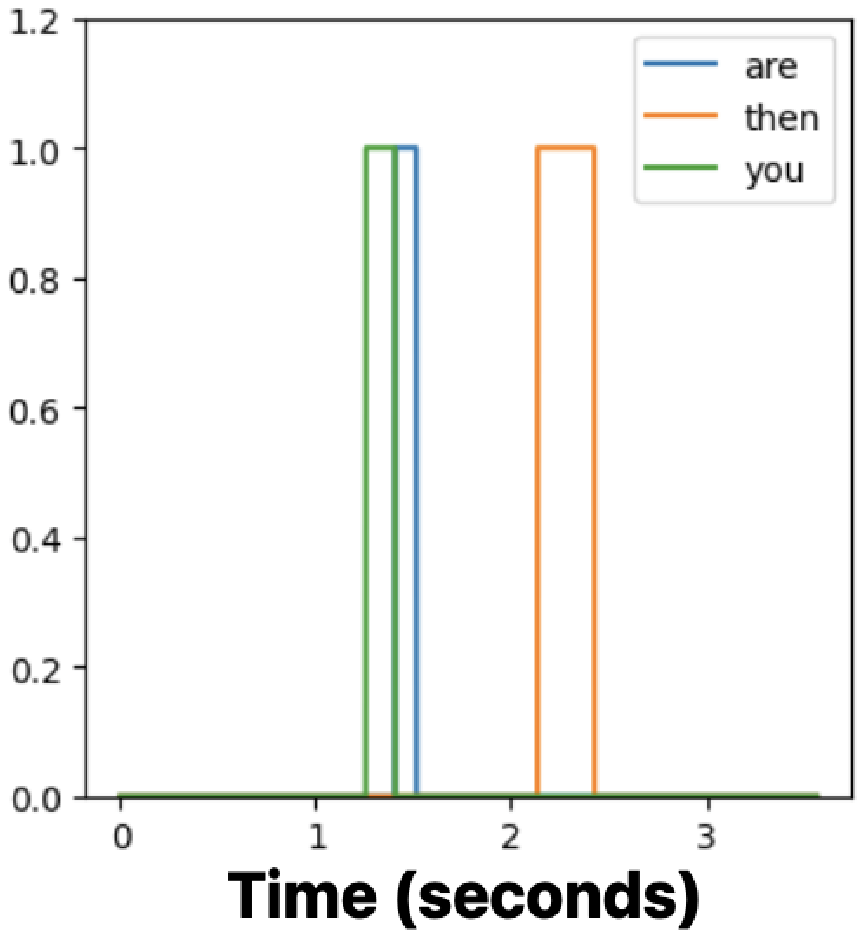}
        \caption{ground-truth events}
    \end{subfigure}
    \begin{subfigure}[b]{0.45\columnwidth}
        \includegraphics[width=0.8\columnwidth]{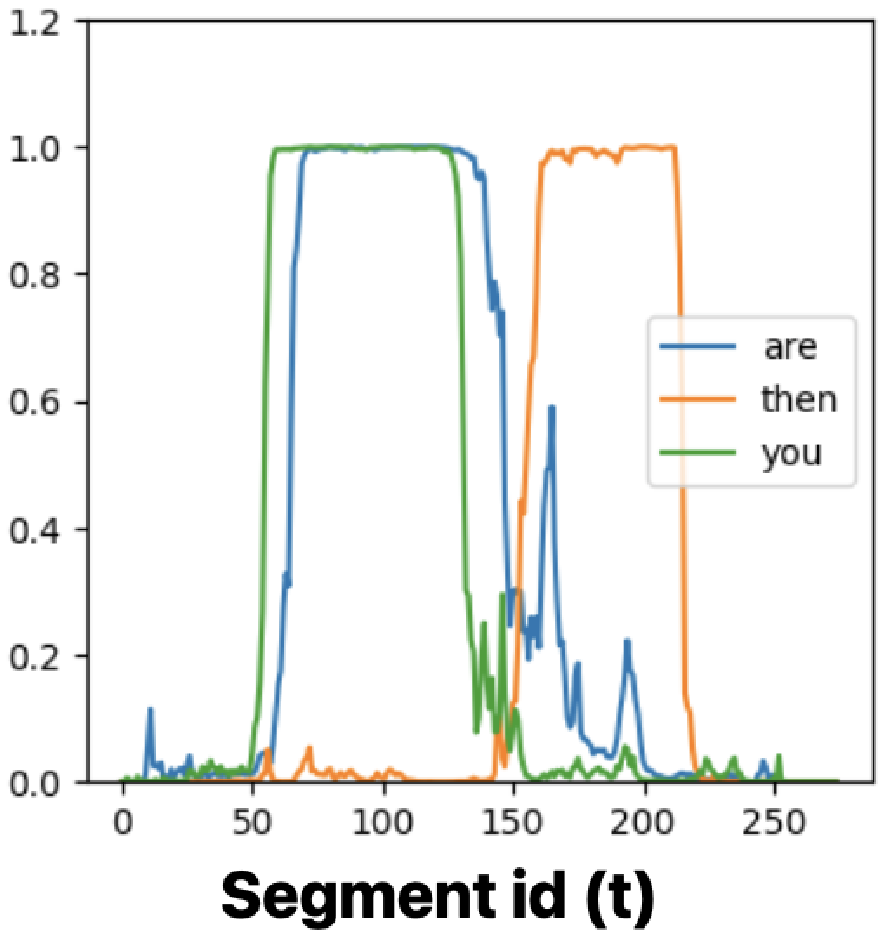}
        \caption{detection signal $\hat{y}$}
    \end{subfigure}
    \begin{subfigure}[b]{0.45\columnwidth}
        \includegraphics[width=0.8\columnwidth]{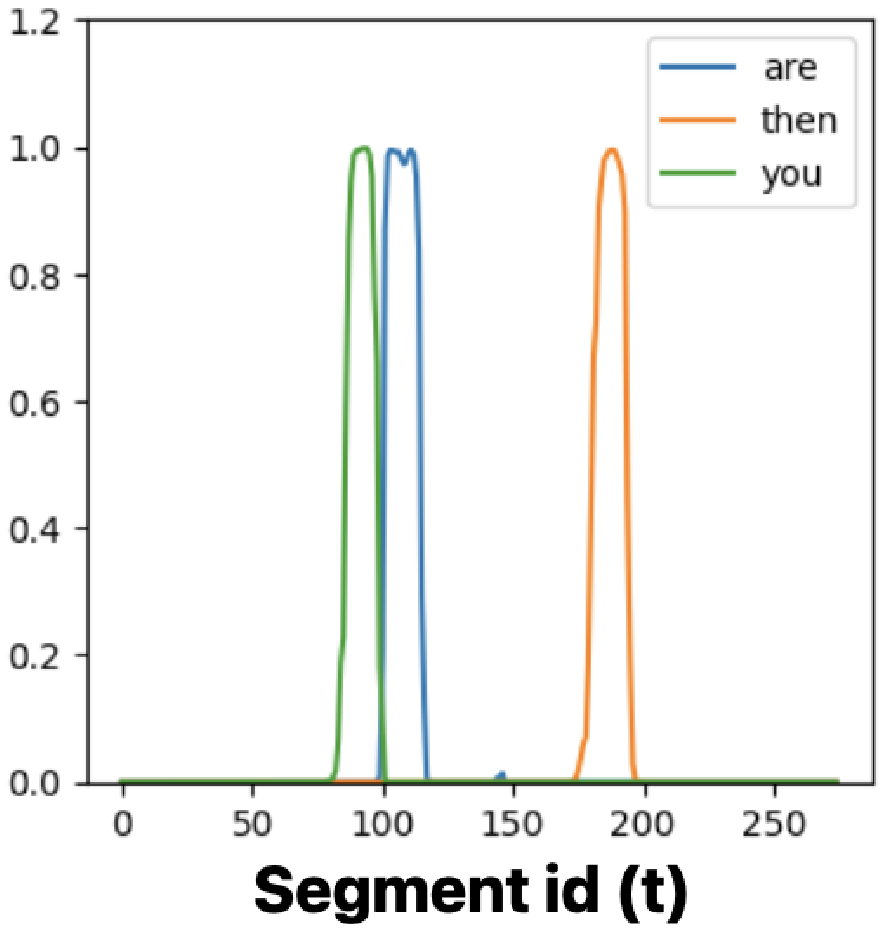}
        \caption{classification result $\hat{s}$}
    \end{subfigure}
    \begin{subfigure}[b]{0.45\columnwidth}
        \includegraphics[width=0.8\columnwidth]{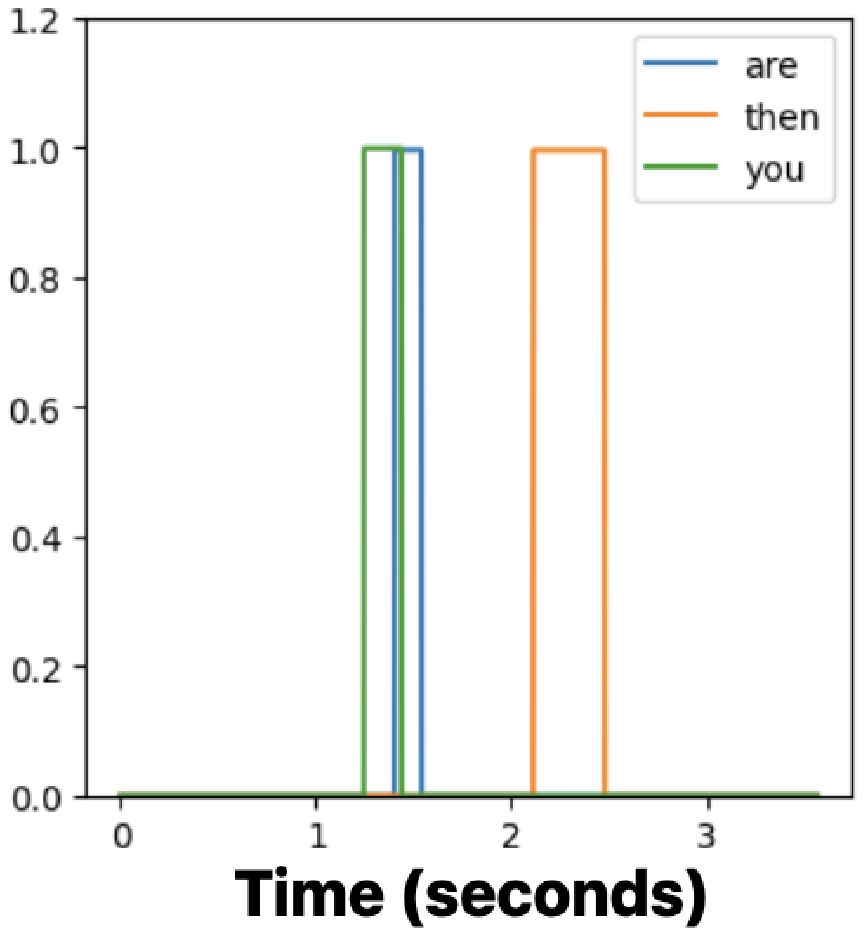}
        \caption{NMS proposed events}
    \end{subfigure}
    
    \caption{Example input and corresponding outputs.}
    \label{fig:filter}
    \vspace{-0.5cm}
\end{figure}

\noindent{\bf Utterance processing (inference only).} Once the model is trained, we can query it with an audio and receive proposed events.
Figure~\ref{fig:filter}-(a) shows the ground-truth events for an example audio, and the non-zero columns of the predicted $\hat{y}$ are plotted in Figure~\ref{fig:filter}-(b). 
The detection $\hat{y}$ is then used in the classifier (Eq.~\ref{eq:softmax}) to compute $\hat{s}$, illustrated in Figure~\ref{fig:filter}-(c). 
At frame $t$, if the maximum prediction $\max(\hat{s}_t)$ is larger than some threshold $\lambda$, an event is proposed as: 
\begin{equation}
\begin{aligned}
 \{event: (\argmax(\hat{s}_t), \hat{b}_t,\  \hat{e}_t), score: \max(\hat{s}_t)\}\\
 \hat{b}_t = S(c_t + \hat{o}_t) - \frac{\hat{l}_t}{2}R,\qquad \hat{e}_t = \hat{b}_t + \hat{l}_tR
\end{aligned}
\end{equation}
where $\hat{b}_t$ and $\hat{e}_t$ are the estimated event begin and event end times.
The extracted events are a set of overlapping windows in the time domain, each of which has a score. To suppress repetitive proposals, we use NMS. The NMS output is illustrated in Figure~\ref{fig:filter}-(d); as seen, the proposed events are quite accurate compared to the ground-truth events.




\section{Evaluations}
\begin{table}[h]
\centering
\caption{Model architecture used in our experiments. Here, ``dim'' is layer output dimension (ignoring batch dimension, assuming the input audio length is equal to the receptive field), ``k'' is kernels size, ``s'' is stride, and ``d'' is dilation.}\label{tab:bcresnet}
\resizebox{\columnwidth}{!}{
\begin{tabular}{cccccc}
\textbf{}                                                 & \textbf{Module} & \textbf{dim}      & \textbf{k} & \textbf{s} & \textbf{d} \\ \hline
\multicolumn{1}{c|}{\textbf{Input}}                       & -               & {[}13200{]}       & -          & -          & -          \\ \hline
\multicolumn{1}{c|}{\textbf{Feature Extractor}}           & FBank           & {[}1, 40, 81{]}   & 400        & 160        & 1          \\ \hline
\multicolumn{1}{c|}{\multirow{12}{*}{\textbf{BC-ResNet}}} & Conv2d          & {[}256, 20, 77{]} & (5, 5)     & (2,1)      & (1,1)      \\
\multicolumn{1}{c|}{}                                     & Transition      & {[}128, 20, 75{]} & 3          & (1,1)      & (1,1)      \\
\multicolumn{1}{c|}{}                                     & Normal          & {[}128, 20, 73{]} & 3          & (1,1)      & (1,1)      \\
\multicolumn{1}{c|}{}                                     & Transition      & {[}192, 10, 69{]} & 3          & (2,1)      & (1,2)      \\
\multicolumn{1}{c|}{}                                     & Normal          & {[}192, 10, 65{]} & 3          & (1,1)      & (1,2)      \\
\multicolumn{1}{c|}{}                                     & Transition      & {[}256, 5, 57{]}  & 3          & (2,1)      & (1,4)      \\
\multicolumn{1}{c|}{}                                     & Normal          & {[}256, 5, 49{]}  & 3          & (1,1)      & (1,4)      \\
\multicolumn{1}{c|}{}                                     & Normal          & {[}256, 5, 41{]}  & 3          & (1,1)      & (1,4)      \\
\multicolumn{1}{c|}{}                                     & Normal          & {[}256, 5, 33{]}  & 3          & (1,1)      & (1,4)      \\
\multicolumn{1}{c|}{}                                     & Transition      & {[}320, 5, 17{]}  & 3          & (1,1)      & (1,8)      \\
\multicolumn{1}{c|}{}                                     & Normal          & {[}320, 5, 1{]}   & 3          & (1,1)      & (1,8)      \\
\multicolumn{1}{c|}{}                                     & Conv2d          & {[}128, 1, 1{]}   & (5, 1)     & (1,1)      & (1,1)      \\ \hline
\multicolumn{1}{c|}{\textbf{Feature (z)}}                 & -               & {[}1, 128{]}      & -          & -          & -          \\ \hline
\end{tabular}
}
\vspace{-0.5cm}
\end{table}

\noindent{\bf Architecture.} We use the BCResNet architecture~\cite{kim2021broadcasted} to implement the backbone in Figure~\ref{fig:overall}. The input to the model is raw audio sampled at $16Khz$. The receptive field is $R=13200$ ($825$ ms) and stride is $S=160$ ($10$ ms). The layers and output dimensions per layer are shown in Table~\ref{tab:bcresnet} when the raw audio length is equal to R. 
The entire CNN backbone encodes each segment $X_t$ as a 128-dimensional vector.

\noindent{\bf Dataset.} Similar to prior work~\cite{segal2019speechyolo}, we use the Montreal Forced Aligner to extract the start and end time of words in LibriSpeech dataset. The lexicon is similarly chosen as 1000 words that appear the most in the training set. 

\noindent{\bf Training.} We train the network for 100 epochs with the Adam optimizer~\cite{kingma2014adam} and a Cosine Annealing scheduler that starts from a learning rate of $0.001$ and gradually reduces it to $0.0001$ by the end of training. During training, we randomly cut a small portion from the beginning f each utterance to make the model agnostic to shifts. No other data augmentation is applied during training. The batch size in our experiments is 32 per GPU, and we use Pytorch distributed data parallelism to train our model on 8 NVIDIA-V100 GPUs.

\subsection{Word detection and localization}
For evaluation, we run NMS on events that have $score \geq \lambda$ to obtain the proposed events. 
We then compute true positives (TP) as the number of proposed events that overlap with a ground-truth event of the same class. If a ground-truth event is not predicted, we count it as a false negative (FN). If a proposed event is not in the ground truth events or the predicted class is wrong we count the event as a false positive. We then compute precision, recall, F1-score, actual accuracy, and average IOU the same way SpeechYolo does~\cite{segal2019speechyolo}. Table~\ref{tab:comparison} summarizes the performance comparison between our method and SpeechYolo. Here, ``Ours-L'' represents the model in Table~\ref{tab:bcresnet} and ``Ours-S'' is the same network with half number of features per layer. We report the SpeechYolo performance numbers from the original paper\footnote{We were not able to reproduce their results as the authors do not provide pre-processed inputs in their github repository}. Our large model is 94\% smaller than SpeechYolo. It can process arbitrary-long audios compared to 1-second segments in SpeechYolo. Our F1-score, IOU, and actual accuracy are also consistently higher.

\begin{table}[ht]
\centering
\caption{Comparison of our trained models with SpeechYolo. In each column, metrics that outperform SpeechYolo are bold. The decision threshold $\lambda$ is separately tuned for SpeechYolo and our work ($\lambda=0.95$) to maximize the F1-scores.}\label{tab:comparison}
 \setlength{\tabcolsep}{5pt}
\resizebox{\columnwidth}{!}{
\begin{tabular}{cccccccc}
\hline
\textbf{Data}                                     & \textbf{Method} & \textbf{\begin{tabular}[c]{@{}c@{}}Model \\ Size\end{tabular}} & \textbf{Prec.} & \textbf{Recall} & \textbf{F1}    & \textbf{\begin{tabular}[c]{@{}c@{}}Actual\\ Acc.\end{tabular}} & \textbf{IOU}   \\ \hline
\multicolumn{1}{c|}{\multirow{3}{*}{test\_clean}} & SY~\cite{segal2019speechyolo}      & 108 MB                                                         & 0.836          & 0.779           & 0.807          & 0.774                                                          & 0.843          \\
\multicolumn{1}{c|}{}                             & Ours-L            & \textbf{6.2 MB}                                                & \textbf{0.863} & \textbf{0.880}  & \textbf{0.872} & \textbf{0.873}                                                 & \textbf{0.857} \\
\multicolumn{1}{c|}{}                             & Ours-S            & \textbf{2.1 MB}                                                & \textbf{0.852} & 0.770           & \textbf{0.809} & 0.759                                                          & \textbf{0.855} \\ \hline
\multicolumn{1}{c|}{\multirow{3}{*}{test\_other}} & SY~\cite{segal2019speechyolo}      & 108 MB                                                         & 0.697          & 0.553           & 0.617          & -                                                              & -              \\
\multicolumn{1}{c|}{}                             & Ours-L            & \textbf{6.2 MB}                                                & \textbf{0.764} & \textbf{0.713}  & \textbf{0.738} & \textbf{0.704}                                                 & \textbf{0.850} \\
\multicolumn{1}{c|}{}                             & Ours-S            & \textbf{2.1 MB}                                                & \textbf{0.777} & 0.549           & \textbf{0.643} & \textbf{0.531 }                                                         & \textbf{0.849}          \\ \hline
\end{tabular}
}
\end{table}

The better performance of our work is due to the fact that we customize the underlying detector design (Figure~\ref{fig:overall}) specifically to overcome challenges in acoustic modeling. To illustrate the effect of different aspects of our design, we perform an ablation study with the large model in Table~\ref{tab:ablation}. In summary, $\hat{s}$ improves precision, $\hat{y}$ improves recall, and $(\hat{o}, \hat{l})$ improve the localization capability (IOU).

\begin{table}[]
\centering
\caption{Effect of the predicted signals on performance. }\label{tab:ablation}
\resizebox{0.7\columnwidth}{!}{
\begin{tabular}{cccccc}
$\mathbf{\hat{y}}$   & $\mathbf{\hat{s}}$   & ($\mathbf{\hat{o}}, \mathbf{\hat{l}}$) & \textbf{Precision} & \textbf{Recall} & \textbf{IOU} \\ \hline
    Yes & No  & Yes     &  0.338         & 0.871       & 0.807    \\
     No  & Yes & Yes     & 0.835         &  0.652      &  0.848   \\
      Yes & Yes & No      & 0.892          &  0.522      & 0.396    \\
      Yes & Yes & Yes     & 0.863          &  0.880      &  0.857   \\ \hline
\end{tabular}
}
\end{table}

\subsection{keyword spotting}
In the next analysis, we compare our method in keyword spotting where the goal is to accurately identify a limited set of words, i.e., 20 words defined in Table~2 of ~\cite{palaz2016jointly} and used by speechYolo. The evacuation metric here is  Term Weight Value (TWV)~\cite{fiscus2007results} defined as follows:
\begin{equation}
    TWV(\lambda) = 1 - \frac{1}{K}[P_{miss}(k, \lambda) + \beta P_{FA}(k, \lambda)]
\end{equation}
where $k\in\{1, \dots, K\}$ refers to the keywords, $P_{miss}(k, \lambda)$ is the probability of missing the $k$-th keyword, and $P_{FA}(k, \lambda)$ is the probability of falsely detecting the $k$-th keyword in 1 second of audio data. Here, $\beta=999.9$ is a constant that severely penalizes the TWV score for high false alarm rates.  
Table~\ref{tab:kws_comparison} compares our keyword spotting performance with SpeechYolo, where the $\lambda$ is tuned for each keyword such that the maximum TWV value (MTWV) is achieved. As seen, our method outperforms SpeechYolo.

\begin{table}[h]
\centering
\caption{MTWV scores for 20-keyword spotting application. Bold numbers are the ones that outperform SpeechYolo. }\label{tab:kws_comparison}
\begin{tabular}{cccc}
Partition   & SY~\cite{segal2019speechyolo} & Ours-L & Ours-S \\ \hline
test\_clean &  0.74          &  \textbf{0.80}   &  0.74\\
test\_other &  0.38          &  \textbf{0.64}   & \textbf{0.53}\\ \hline
\end{tabular}
\end{table}
\vspace{-0.5cm}

\section{Conclusion}
This paper proposes a solution to limited-vocabulary word detection and localization in speech data. We devise model components that ensure a high precision, recall, and localization score. We then define the loss functions required to train the model components. We showed in our experiments that, compared to existing work, our model is more accurate, smaller in size, and capable of processing arbitrary length audio.

\bibliographystyle{IEEEbib}
\bibliography{refs}

\end{document}